\documentclass[conference]{IEEEtran}
\IEEEoverridecommandlockouts
\usepackage{cite}
\usepackage{sidecap}
\usepackage{amsmath,amssymb,amsfonts}
\usepackage{rotating}
\usepackage{url}
\usepackage{algorithmic}
\usepackage{graphicx}
\usepackage{textcomp}
\usepackage{multirow}
\usepackage{multicol}
\usepackage{comment}
\usepackage{balance}
\usepackage[table]{xcolor}
\def\BibTeX{{\rm B\kern-.05em{\sc i\kern-.025em b}\kern-.08em
    T\kern-.1667em\lower.7ex\hbox{E}\kern-.125emX}}
\begin{document}
\title{Resetting the baseline:  CT-based COVID-19 diagnosis with Deep Transfer Learning is not as accurate as widely  thought\\
}



\author{\IEEEauthorblockN{Fouzia Altaf}
\IEEEauthorblockA{\textit{School of Science} \\
\textit{Edith Cowan University}\\
Joondalup, Australia \\
faltaf@our.ecu.edu.au}

\and
\IEEEauthorblockN{ Syed M.S. Islam}
\IEEEauthorblockA{\textit{School of Science} \\
\textit{Edith Cowan University}\\
Joondalup, Australia \\
syed.islam@ecu.edu.au}
\and
\IEEEauthorblockN{Naveed Akhtar}
\IEEEauthorblockA{\textit{
Computer Science \& Software Engineering} \\
\textit{University of Western Australia}\\
Crawley, Australia \\
naveed.akhtar@uwa.edu.au}
}

\maketitle

\begin{abstract}
Deep learning is gaining instant  popularity in computer aided diagnosis of COVID-19. Due to the high sensitivity of Computed  Tomography (CT) to this disease, CT-based COVID-19 detection with visual models is currently at the forefront of medical imaging research. Outcomes published in this direction are frequently claiming highly accurate detection under deep transfer learning. This is leading medical technologists to believe that deep transfer learning is the mainstream solution for the problem. However, our critical analysis of the literature reveals an alarming performance disparity between different published results. Hence, we conduct a  systematic thorough investigation to analyze the effectiveness of deep transfer learning for COVID-19 detection with CT images. Exploring 14 state-of-the-art visual models with over 200 model training  sessions, we conclusively establish that the published literature is frequently overestimating transfer learning performance for the problem, even in the  prestigious scientific sources. The roots of overestimation trace back to inappropriate data curation. We also provide case studies that consider more realistic scenarios, and establish transparent baselines for the problem.  We hope that our reproducible investigation will help in curbing hype-driven claims for the critical problem of COVID-19 diagnosis, and pave the way for a more transparent performance evaluation of techniques for CT-based COVID-19 detection.
\end{abstract}


\section{Introduction}
Likely originated in Wuhan, China in December 2019~\cite{wang2020clinical}, a novel coronavirus disease, later dubbed COVID-19~\cite{WHO1} was  declared a pandemic in March 2020~\cite{WHO2}. Since then, it has disrupted nearly all aspects of daily life across the globe. The onus is now on  the scientific community to rapidly develop solutions to curb this disease. This has spawned numerous interdisciplinary efforts in multiple scientific fields. Whereas cross-domain research is essential to meet the challenges of COVID-19, it can also sprout misleading findings that lack in accounting for detailed knowledge of the contributing domains. 
This concern is especially valid for COVID-19 research that is driven by the highest level of urgency which can cause superficial investigations.  
This paper exposes a scenario marred by this issue in medical image analysis literature for COVID-19 detection using CT images. 
It then addresses the concern by providing an extensive transparent investigation. 


The Reverse
Transcription Polymerase Chain Reaction (RT-PCR) is currently considered the gold standard for diagnosing COVID-19. Nevertheless, it can still be complemented for improved detection with imaging techniques, e.g.~radiography, tomography. The later can also be a viable diagnostic tool in the absence of RT-PCR. It is claimed that Computed Tomography (CT) has even higher sensitivity to COVID-19 than RT-PCR~\cite{wynants2020prediction, long2020diagnosis, fang2020sensitivity}. This has led to a significant interest of medical imaging community in exploring CT images for COVID-19 detection \cite{ardakani2020application, ko2020covid, pham2020comprehensive, pathak2020deep, gozes2020rapid, altaf2021boosting}. Medical imaging researchers currently rely strongly on the developments in the fields of machine learning and computer vision~\cite{altaf2019going}, where deep learning~\cite{lecun2015deep} is the key technology that is providing numerous breakthroughs. This has invited medical researchers to develop deep learning solutions for CT-based COVID-19 detection.

Unfortunately, deep learning requires a large amount of annotated (training) data to learn useful computational models to accurately detect COVID-19 using CT images. Such a large volume of data, which is annotated by medical experts, is currently not available for this task. This makes `deep transfer learning' as the preferred solution for the problem~\cite{altaf2019going}. Put simply, transfer learning allows one to transfer a deep learning model trained for one domain (e.g.~natural images) to another domain (e.g.~CT images). Due to the easy availability of models trained on large-scale annotated natural images, it is a common practice to transfer natural image models to the domain of CT images for our concerned problem. This broad strategy is currently highly popular in the related literature, however, we also find it marked  with confusing results. 

In the current literature, at one end we find contributions that claim highly  precise detection of COVID-19 with transfer learning using very limited amount of annotated training data~\cite{pham2020comprehensive}, \cite{ahuja2021deep},  \cite{shah2021diagnosis}. These contributions make their claims based on vanilla transfer learning. Contrastingly, there are also works that enhance  transfer learning for better performance with rather sophisticated procedures. Surprisingly, most of such methods do not claim very  high accuracies despite their enhancements~\cite{altaf2021boosting}, \cite{loey2020deep}, \cite{gozes2020rapid}.
On the other extreme, we also find contributions that train models from scratch~\cite{rahimzadeh2021fully}, \cite{yang2020deep}, \cite{li2020artificial}. When large amount of training data is available, this strategy is expected to give the best performance. However, we often find methods in this category reporting comparable~\cite{rahimzadeh2021fully}, \cite{li2020artificial}, \cite{yang2020deep} or lower \cite{song2020end}, \cite{wu2020deep} performance than the vanilla transfer learning methods. These method often use orders of magnitude larger data as compared to transfer learning methods, with few exceptions, e.g.~\cite{yang2020deep}, that  claim high performance with small datasets. We capture a summary of this contradiction in the published literature with representative examples in Table.~\ref{tab:literature}.

\begin{table*}
    \centering
    \caption{Contrasting results for CT-based COVID-19 diagnosis with deep learning are found in the literature. Transfer learning methods generally claim high performance with small training data. Enhancements of transfer learning (Transfer + other) often claim lower performance. Training `From scratch' is also frequently reported  to have lower performance than transfer learning methods despite utilizing large number of images. The mentioned number of images is for full dataset (training +testing). Refer to Section~\ref{sec:Rb} for definitions of the used metrics.}
    \begin{tabular}{c|c|c|c|c|c|c|c}
    \hline
    \multirow{2}{*}{\textit{\textbf{Reference}}} & 
    \multirow{2}{*}{\textit{\textbf{Model/architecture}}}&
    \multirow{2}{*}{\textit{\textbf{No. of images}}}& \multicolumn{4}{c|}{\textit{\textbf{Results}}} & \multirow{2}{*}{\textit{\textbf{Training}}}\\ \cline{4-7} 
     &&& Acc (\%). & Sens. (\%) & Spec. (\%) & F1-score \\ \hline
     Ardakani et al.~\cite{ardakani2020application} & Xception & 1020 & 100 & 98.04 & 100 & - & Transfer learning \\
     Ahuja et al.~\cite{ahuja2021deep}& ResNet-18&406 & 99.4 & 100 & 98.6 & 0.99 &  Transfer learning \\ 
     Ko et al.~\cite{ko2020covid}& ResNet-50 &3993 &99.8 & 99.5 & 100 & - & Transfer learning\\
     Pham~\cite{pham2020comprehensive}& DenseNet-201 & 746 & 96.2 & 95.7 & 96.6 & 0.96 &  Transfer learning \\
     Shah et al.~\cite{shah2021diagnosis}& VGG-19&738 & 94.5 & - & - & - & Transfer learning  \\
     Pathak et al.~\cite{pathak2020deep}& ResNet-50&852 & 93.0 & 91.45 & 94.77 & - &  Transfer learning\\  \hline
     Gozes et al.~\cite{gozes2020rapid}& ResNet-50&1865 & - & 98.2 & 92.2 & - & Transfer + other \\ 
     Loey et al.~\cite{loey2020deep}& ResNet-50&742 & 82.9 & 77.66 & 87.62 & -  & Transfer + other \\
     Altaf et al.~\cite{altaf2021boosting}  & Inception-v3 & 746 & 78.5 & 88.6 & 67.1 & 0.81 & Transfer + other \\\hline
    
     Rahimzadeh~\cite{rahimzadeh2021fully}& ResNet-50& 48260 & 98.5 & - & - & - & From scratch\\
     Yang et al.~\cite{yang2020deep}& DenseNet & 295 & 95 & 100 & 90 & 0.95 & From scratch\\
     
     Li et al.~\cite{li2020artificial}& ResNet-50-based &4563 &- & 90.0 & 96.0 & - & From scratch\\
     Song et al.~\cite{song2021deep}& ResNet-50-based &1993 &86.0 & 96.0 & 77.0 & 0.87 & From scratch \\
     Wu et al.~\cite{wu2020deep}& ResNet-50-based &495 & 76.0 & 81.1 & 61.1 & - & From scratch \\
     \hline
    \end{tabular}
        \label{tab:literature}
\end{table*}

Contrasting results in the literature call for an in-depth investigation that can provide a transparent baseline for the critical task of COVID-19 detection with CT imaging - which is being widely preferred for its high sensitivity to COVID-19~\cite{wynants2020prediction}, \cite{long2020diagnosis}, \cite{fang2020sensitivity}. Such an investigation for deep transfer learning is vital for two main reasons. First, it will halt premature deployment of this technology in practice, which is currently encouraged by  frequent overestimation of its capabilities for the problem at hand. Second, a transparent perspective on deep transfer learning abilities for CT-based COVID-19 detection will allow the scientific community to better address weaknesses of this strategy, which are currently not apparent due to the high performance claims. 

In this paper, we address this issue with a reproducible comprehensive study that resets the baseline of deep transfer learning for CT-based COVID-19 detection. To that end, we extensively analyse performance of 14 state-of-the-art models, pre-trained on ImageNet~\cite{deng2009imagenet} - a large-scale database of natural images, transferred to CT image domain with COVID-CT-Dataset (CCD)~\cite{zhao2020covid}. Our choice of CCD is based on the recent comprehensive study published in Nature Scientific Reports~\cite{pham2020comprehensive}. Our analysis is also mainly anchored by the experimental setup of \cite{pham2020comprehensive}. However, we find that their evaluation does not strictly follow the practices of machine learning community. Doing so results in a considerable reduction in the claimed transfer learning performance estimates\footnote{We emphasize that our study is not meant to invalidate  \cite{pham2020comprehensive}. Using \cite{pham2020comprehensive}, we mainly expose a typical cause of performance overestimation in this interdisciplinary research direction that combines technologies from machine learning, medical imaging and computer vision. We thank the authors of \cite{pham2020comprehensive} for providing their implementation.}. Our investigation identifies cursory data curation as the major cause of a drastic overestimation of performance for the problem in the broader literature. We establish a transparent baseline for CCD, and  also analyse two practicable scenarios to enable a more informed expectation of deep transfer learning based COVID-19 detection with CT images in practice. 

The contributions of this paper are summarized as follows:
\begin{itemize}
    \item We identify the overestimation of transfer learning abilities in the literature for CT-based COVID-19 detection.
    \item We perform a transparent investigation to reset the baseline of COVID-19 detection using transfer learning on CCD~\cite{zhao2020covid}. We demonstrate the performance overestimation by comparing results with a study published on the same topic in Nature Scientific Reports~\cite{pham2020comprehensive}.
    \item We establish inappropriate data curation as the source of overestimated results.
    \item We also provide transfer learning results for COVID-19 detection under  pragmatic scenarios.
\end{itemize}

\section{Background and Related Work}
To comprehend the results and claims in the literature on CT-based COVID-19 detection with transfer learning, it is imperative to understand the interaction between the involved scientific sub-fields and the metrics used in evaluation. Before discussing the related works, we first provide a short discussion on these topics.

\vspace{1mm}
\noindent{\bf Background:} Deep learning~\cite{lecun2015deep} is the key technology around which imaging based COVID-19 research currently revolves. In itself, its a representation learning technique, which is a core topic in the field of machine learning~\cite{bengio2013representation}. However, computer vision researchers have also found deep learning with Convolutional Neural Networks (CNNs) to be extremely effective for their   problems~\cite{krizhevsky2012imagenet}. From computer vision literature, CNNs paved their way to medical image analysis~\cite{altaf2019going}. Medical image analysis is already a domain where experts in medical science and image processing interact. 

Currently, the claims of highly accurate computer aided diagnosis of COVID-19 are based on deep learning techniques, mainly found in the medical imaging literature.
As clear from the discussion above, deep learning is a tool explored in multiple domains. Nearly, all of those domains claim it to be highly accurate~\cite{lecun2015deep}. When this tool is used by non-experts of machine learning (the originating field  of modern deep learning) in scenarios dictated by urgency, hype-drive  overestimation of its capabilities are highly likely. Our analysis in the subsequent sections reveals that this phenomenon has also affected CT-based COVID-19 diagnosis research. 

In the text to follow and Table~\ref{tab:literature}, we refer to different evaluation metrics to discuss performance. For the convenience of readers, we provide formal definitions of these metrics upfront, and follow them throughout the paper.
For the considered binary classification problem, let us denote true positive predictions by TP, true negatives by TN, false positives by FP and false negatives  by FN. The Specificity (Spec.),   Sensitivity (Sens.) and F1-Score are then computed as:
\begin{align*}
    \text{Spec.} = \frac{\text{TN}}{\text{TN} + \text{FP}} \times 100\% 
    ~,~\text{Sens.} = \frac{\text{TP}}{\text{TP + FN}} \times 100\%,
    \end{align*}
    \begin{align*}
    \text{F1-Score} = 2 \times \frac{\text{PPV} \times \text{TPR}} {\text{PPV} + \text{TPR}},
\end{align*}
where PPV $=$ TP/(TP+FP) and TPR $=$ TP/(TP+FN). Based on these metrics, we compute Accuracy (Acc.) as
\begin{align*}
    \text{Acc.} = \frac{\text{TP} +\text{TN}}{\text{TP} +\text{TN} + \text{FP} + \text{FN}} \times 100\%.
\end{align*}

\vspace{3mm}
\noindent{\bf{Related Work:}}
Since the outbreak of COVID-19, numerous  works have employed deep learning for  detection and classification of COVID-19  from CT images. For instance, Zheng et al.~\cite{zheng2020deep}  proposed a weakly supervised deep learning method using 3D CT volumes for the detection of COVID-19. The model is trained on 499 CT volumes and tested on 131 CT volumes. Chen et al.~\cite{chen2020deep} constructed a deep learning based system for the diagnosis of COVID-19 and phenumonia from CT images. 
Trained with private images (in access of 46K), their method is claimed to have achieved $98.85\%$ accuracy.
Yousefzadeh et al.~\cite{yousefzadeh2021ai} proposed a deep learning framework ai-corona  to assist radiologist in diagnosing COVID-19 in CT images. In their experiments,  they used more than five thousand CT scans. Mei et al.~\cite{mei2020artificial} claim that their model outperformed radiologists in diagnosing COVID-19 positive patients in CT scans. Their approach has three stages including CNN followed by SVM and multi-layer perceptrons. They also combine other clinical information in their predictions. 

More closely related to our study are the works that fully focus on transfer learning for CT-based COVID-19 diagnosis \cite{ardakani2020application}, \cite{ko2020covid}, \cite{pham2020comprehensive}, \cite{pathak2020deep}, \cite{ahuja2021deep},   \cite{shah2021diagnosis}. The main goal of such contributions is to often empirically establish the best-performing state-of-the-art visual models for transfer learning on CT images for COVID-19. As identified in Table~\ref{tab:literature}, so far, these works have claimed a variety of popular visual models, e.g.~Xeception, ResNets, DenseNet, VGG-19, as the top-performing models. We also frequently witness very high performance reported by these methods while using very limited training data. Of particular relevance to our work is the  study presented in \cite{pham2020comprehensive}. We closely follow \cite{pham2020comprehensive} in our evaluations, however we provide very different results that demystify the claims of high performance of transfer  learning in the context of CT-based COVID-19 diagnosis. Besides the above-mentioned literature,  other works are also appearing in this direction. We refer interested readers to \cite{roberts2021common} for a recent survey.

\begin{sidewaystable}
{\small
\setlength{\tabcolsep}{4pt} 
\renewcommand{\arraystretch}{1}
   \caption{Quantitative proof of overestimation of deep transfer learning performance on CCD~\cite{zhao2020covid}. `Actual' results are computed following the standard five-fold cross-validation protocol in machine learning. `Pham~\cite{pham2020comprehensive}' perform 80-20\% training-validation data splitting five times by sampling validation data uniformly at random from CCD~\cite{zhao2020covid}. The `Diff.' is the difference between mean values of `Pham' and `Actual'. Larger values of `Diff.' indicate larger overestimation. The largest `Diff.' in each case is highlighted in red, the smallest is highlighted in gray. Largest values each of the remaining column is bold-faced.}
  \begin{tabular}{l|c|c|c|c|c|c|c|c|c|c|c|c}
    \hline
    \multirow{2}{*}{\textit{\textbf{Models}}} &
      \multicolumn{3}{c|}{\textit{\textbf{Accuracy (\%)}}} & 
      \multicolumn{3}{c|}{\textit{\textbf{Sensitivity (\%)}}} &
      \multicolumn{3}{c|}{\textit{\textbf{Specificity (\%)}}}&
      \multicolumn{3}{c}{\textit{\textbf{F1-score}}} \\ \cline{2-13}
    & Actual & Pham~\cite{pham2020comprehensive} &  \cellcolor{yellow!18}Diff. & Actual & Pham~\cite{pham2020comprehensive} & \cellcolor{yellow!18}Diff. & Actual & Pham~\cite{pham2020comprehensive} & \cellcolor{yellow!18}Diff. & Actual & Pham~\cite{pham2020comprehensive} & \cellcolor{yellow!18}Diff. \\
    \hline
    GoogLeNet & 68.00 $\pm$10.93  & 93.83 $\pm$ 6.97 & \cellcolor{red!18}25.83 & 74.72$\pm$ 22.03 & 96.71 $\pm$ 4.06 & 21.99 & 60.45$\pm$15.97 & 90.57 $\pm$ 10.53 & \cellcolor{red!18}30.12 & 0.70$\pm$0.12 & 0.94 $\pm$ 0.06 & 0.24 \\
    SqueezeNet & 73.06 $\pm$ 5.69 & 87.52 $\pm$ 6.45 & 14.46 & \textbf{81.23$\pm$ 20.32} & 86.84 $\pm$ 10.11 & \cellcolor{black!15}5.61 & 63.85$\pm$15.24 & 88.29 $\pm$ 12.01 & 24.44 & 0.75$\pm$0.19 & 0.88 $\pm$ 0.06 & \cellcolor{black!15}0.13 \\
    ShuffleNet & 71.21 $\pm$ 6.18 & 95.97 $\pm$ 5.09 & 24.76 & 74.21$\pm$17.38 & 95.44 $\pm$ 7.47 & 21.23 & 67.87$\pm$14.89 & 96.57 $\pm$ 2.96 & 28.70 & 0.72$\pm$0.08 & \textbf{0.96 $\pm$ 0.05} & 0.24 \\
    ResNet-18 & 70.59 $\pm$ 12.41 & 95.44 $\pm$ 8.02 & 24.85 & 62.22$\pm$30.44 & \textbf{98.99 $\pm$ 1.65} & \cellcolor{red!18}36.77 & 80.21$\pm$13.78 & 91.43 $\pm$ 15.25 & 11.22 & 0.66$\pm$0.18 & 0.96 $\pm$ 0.07 & \cellcolor{red!18}0.30 \\
    ResNet-50 & 73.11 $\pm$ 10.91 & 93.62 $\pm$ 6.17 & 20.51 & 73.75$\pm$22.59 & 95.57 $\pm$ 6.27 & 21.82 & 73.35$\pm$12.43 & 91.43 $\pm$ 6.06 & 18.08 & 0.73$\pm$0.13 & 0.94 $\pm$ 0.06 & 0.21 \\
    ResNet-101 & 73.77 $\pm$ 13.94 & 93.29 $\pm$ 5.69  & 19.52 & 79.19$\pm$15.31 & 96.20 $\pm$ 1.79 & 17.01 & 67.63$\pm$18.13 & 90.00 $\pm$ 10.10 & 22.37 & 0.76$\pm$0.12 & 0.94 $\pm$ 0.05 & 0.18 \\
    Xception & 72.53 $\pm$ 11.27 & 91.11 $\pm$ 10.14 & 18.58 & 72.07$\pm$11.58 & 89.56 $\pm$ 12.55 & 17.49 & 73.06$\pm$17.17 & 92.86 $\pm$ 7.80 & 19.80 & 0.73$\pm$0.10 & 0.91 $\pm$ 0.10 & 0.18 \\
    Inception-v3 & 73.48 $\pm$ 6.99 & 93.62 $\pm$ 5.22 & 20.14 & 74.88$\pm$12.60 & 96.20 $\pm$ 0.00 & 21.32 & 71.91$\pm$2.42 & 90.71 $\pm$ 11.11 & 18.80 & 0.74$\pm$0.08 & 0.94 $\pm$ 0.07 & 0.20 \\
    Inception-ResNet-v2 & 74.16 $\pm$ 8.97 & 88.59 $\pm$ 7.59 & \cellcolor{black!15}14.43 & 65.10$\pm$15.76 & 89.24 $\pm$ 2.69 & 24.14 & \textbf{84.50$\pm$5.21} & 87.86 $\pm$ 13.13 & \cellcolor{black!15}3.36 & 0.72$\pm$0.12 & 0.89 $\pm$ 0.07 & 0.17 \\
    VGG-16 & 70.23$\pm$ 8.06 & 89.26 $\pm$ 8.80 & 19.03 & 77.67$\pm$15.42 & 92.83 $\pm$ 6.24 & 15.16 & 61.84$\pm$23.31 & 85.24 $\pm$ 14.45 & 23.40 & 0.73$\pm$0.07 & 0.90 $\pm$ 0.08 & 0.17 \\
    VGG-19 & 69.58 $\pm$ 7.50 & 90.16 $\pm$ 7.72 & 20.58 & 63.16$\pm$25.04 & 87.34 $\pm$ 10.36 & 24.18 & 76.82$\pm$21.58 & 93.33 $\pm$ 5.77 & 16.51 & 0.66$\pm$0.15 & 0.90 $\pm$ 0.08 & 0.24 \\
    DenseNet-201 & \textbf{75.89 $\pm$ 8.50} & \textbf{96.20 $\pm$ 4.95} & 20.31 & 77.53$\pm$13.91 & 95.78 $\pm$ 5.27 & 18.25 & 74.19$\pm$11.23 & \textbf{96.67 $\pm$ 4.59} & 22.48 & \textbf{0.77$\pm$0.09} & \textbf{0.96 $\pm$ 0.05} & 0.19 \\
    MobileNet-v2 & 74.42 $\pm$ 7.36 & 95.97 $\pm$ 7.18 & 21.55 & 76.86$\pm$16.06 & 96.71 $\pm$ 6.04 & 19.85 & 71.64$\pm$9.97 & 95.14 $\pm$ 8.55 & 23.50 & 0.75$\pm$0.08 & 0.96 $\pm$ 0.07  & 0.21 \\
    NasNet-Mobile & 69.97 $\pm$ 4.96 & 89.26 $\pm$ 8.14 & 19.29 & 73.31$\pm$9.01 & 91.56 $\pm$ 5.12 & 18.25 & 66.18$\pm$7.38 & 86.67 $\pm$ 13.27 & 20.49 & 0.75$\pm$0.05 & 0.90 $\pm$ 0.07 & 0.15 \\
    \hline
  \end{tabular}
\label{tab:original}
 }
 \end{sidewaystable}

\section{Overestimation of Transfer Learning abilities}
\label{sec:OTL}
To provide a concrete evidence of overestimation of transfer learning abilities in medical imaging literature for our problem, we focus on \cite{pham2020comprehensive} as the representative existing study. Published in \textit{Nature Scientific Reports}, this work provides a baseline  for transfer learning based COVID-19 detection using CT images of  CCD~\cite{zhao2020covid}. The authors made their code public\footnote{\url{https://drive.google.com/file/d/194ucricsWADWgeE1m7cnPKL-8PjN4oai/view}.}. By splitting CCD samples into five different 80\%-20\% training-validation sets per experiment, they performed 5 experiments per model and provided the validation results. On the unseen validation samples, they reported  up to 96.20\% mean  prediction accuracy of the transferred models. 

We copy the results of Pham~\cite{pham2020comprehensive} from the original paper in Table~\ref{tab:original}. We defer further  discussion on the training and validation data sets of  \cite{pham2020comprehensive} to Section~\ref{sec:source}, where we also explore other aspects of the data. Here, it is sufficient to note that the results are  reported using less than 600 annotated CT images in each training session of a given model. High performance reports in \cite{pham2020comprehensive} with such a limited data seems an excellent prospect for transfer learning because a $\sim100\%$ prediction accuracy appears possible here by using a (relatively) small number of additional training images.  However, surprisingly, when we reproduced the results of \cite{pham2020comprehensive} with the standard five-fold cross-validation protocol, there is a drastic reduction in the claimed performance. The comparison of our results with \cite{pham2020comprehensive} is given in Table~\ref{tab:original}. 

We note that, except for the used data samples, we follow \cite{pham2020comprehensive} in our experiments down to every single detail.
That is, as per \cite{pham2020comprehensive}, we first convert the original images to RGB and resize them to input dimensions of the used CNN. We use stochastic gradient descent with momentum for model   optimization, for which the momentum value is set to 0.9. The gradient threshold method with $\ell_2$-norm is used in the  training. We use a batch size of 10, and train the models for 6 epochs to complete the transfer, with a weight decay of 0.0001. The learning rate is set to a fixed value of 0.0003, and the training and validation samples are shuffled before every epoch.
In all our experiments reported in this paper, we keep this hyper-parameter setting fixed. We explicitly note the exceptions whenever they occur.
In Table~\ref{tab:original}, the only difference between our  experiments and \cite{pham2020comprehensive} is in the used data splits. Instead of a \textit{random} 80\%-20\% split of training-validation data of \cite{pham2020comprehensive}, we perform a more \textit{systematic} split. 

In our data splitting, we first sort the samples in CCD (for both COVID positive and negative sets) by their names used in the dataset. Then,  we remove the first 20\% samples from the data and consider those  as  the validation set, where the remainder is  the training set. This gives us the training and validation sets for the 1$^{\text{st}}$ fold. For the 2$^{\text{nd}}$ fold, we put back the 20\% samples taken out for the 1$^{\text{st}}$ fold, and take out the next 20\% samples for the  validation set, the remainder is the training set. We repeat this to construct the training-validation sets for all the  five folds. This procedure reflects the typical  understanding of the `five-fold' evaluation protocol in  machine learning literature. 

Normally, one would expect the mean performance under the standard five-fold protocol and random five splits to roughly coincide. This is because both protocols evaluate performance on unseen data of the same proportion.  However, as apparent in Table~\ref{tab:original}, this is not the case here. An overestimation of 25.83\% accuracy, 36.77\% sensitivity, 30.12\% specificity and 0.3 F1-score is identifiable in the table. It is emphasized that these values are simple differences of the mean  of  metric scores under the two protocols. Converting them to percentages of the original results  \cite{pham2020comprehensive} will lead to even  higher numbers. 
Also notice that, not only the corresponding performance of the same models differs significantly for the two evaluations, the difference in the best performances across all models for any evaluation metric is remarkably high. Even the minimum differences for the same models (highlighted gray) are not minor - despite being anomalies in most cases.    

\begin{figure*}[t!]
    \centering
    \includegraphics[width = 0.9\textwidth]{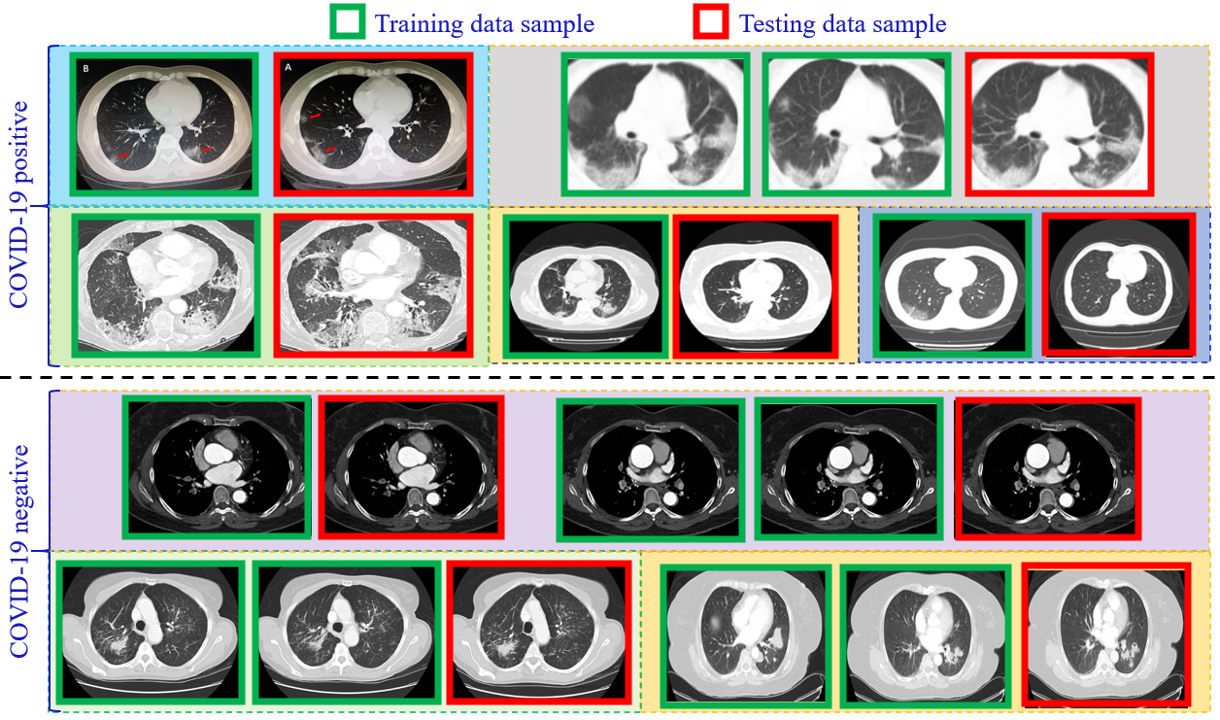}
    \caption{Typical examples of images in CCD~\cite{zhao2020covid}. In both COVID-19 positive (top) and negative (bottom) classes, large visual similarities exist between samples within each class. Representative clusters of such cases are shown, differentiated by different background colors. The visual similarities are mainly along the aspects of color contrast, cropping, slice regions and resemblance of other dominant patterns that are irrelevant to COVID-19 symptoms. These similarities are a by-product of data curation. Sampled uniformly at  `random', the images in green boxes got  chosen as training samples in our experiments, whereas red box images were selected as test data. Due to dominant COVID-19 irrelevant features, individuals having no medical expertise are able to correctly classify the test images without even knowing that these are CT images.}
    \label{fig:samples}
    \vspace{-2mm}
\end{figure*}

\section{The source of overestimation}
\label{sec:source}
The results in Table~\ref{tab:original} provide a clear proof of performance overestimation. This  leads to a natural question that `what really caused such a huge disparity in performance?'. We do not hold the evaluation protocol of  \cite{pham2020comprehensive} responsible for this issue. Provided that images in a dataset are considered representative random  samples of the same image distribution, the average results of random data splitting and sequential data splitting (as in five-fold protocol) can not be expected to differ too  much. Hence, the evaluation protocol of \cite{pham2020comprehensive} is reasonable in general. However, here the issue is with how the  COVID-CT-Dataset (CCD)~\cite{zhao2020covid} is curated.  The dataset  consists of  349 CT images of COVID-19 positive cases and 397 CT images of negative cases. The samples in the  dataset are assembled from different COVID-19 papers published in  medRxive, bioRxive, MedPix, LUNA and PubMed Central (PMC) etc. The curation process often extracts samples directly from the digital copies of the publications themselves. We refer to \cite{zhao2020covid} for the exact details on the extraction process. Here, we  are mainly interested in the nuisance patterns emerging in the dataset due to such a process of sample acquisition. Although, it is  claimed that a senior radiologist from Tongji Hospital, Wuhan, China has confirmed the practicality of this dataset~\cite{zhao2020covid}, our opinion from machine learning viewpoint differs. For the case of learning computational models, this dataset can cause  misleading results, unless carefully (pre-)processed.

To corroborate our claim, we show representative examples of samples in the dataset in Fig.~\ref{fig:samples}. To generate the figure, we picked random samples from a testing (i.e.~validation) set that we created under \textit{random} data splitting of  \cite{pham2020comprehensive}. These images are bounded in red boxes in Fig.~\ref{fig:samples}. 
We then performed a manual selection of green-boxed images from the remaining training set. These images are picked by a human participant by looking at the corresponding red-boxed images. This participant was  given the full training set without labels, and asked to pick the closest match(es) by visual inspection. The  participant was unaware of even the fact that these are CT-images, let alone having any expertise of COVID-19 diagnosis. Interestingly, the participant  picked the corresponding green-box images in each cluster (identified by different background colours in Fig.~\ref{fig:samples}) from the correct classes. It is easy to see that the similarities between red-boxed images and their green-boxed counterparts are hardly COVID-19 relevant. 

\begin{table*}[t!]
    \centering
    \caption{New transfer learning baseline for CCD~\cite{zhao2020covid} under five-fold protocol using data augmentation. Performance gain over transfer learning without data augmentation (Table~\ref{tab:original}) is highlighted green for positive and red for negative values.}
    \begin{tabular}{l||c|c||c|c||c|c||c|c}
    \hline
         \textit{\textbf{Model}} & \textit{\textbf{Accuracy (\%)}} & \textit{\textbf{Gain (\%)}} & \textit{\textbf{Sensitivity (\%)}} & \textit{\textbf{Gain (\%)}} & \textit{\textbf{Specificity (\%)}} & \textit{\textbf{Gain (\%)}} & \textit{\textbf{F1-score}} & \textit{\textbf{Gain (\%)}}\\ \hline
        GoogLeNet & 70.56 $\pm$9.80 & \cellcolor{green!25}3.77 & 77.31$\pm$7.44 & \cellcolor{green!25}3.47 & 62.82$\pm$17.34 & \cellcolor{green!25}3.91 & 0.73$\pm$0.07 & \cellcolor{green!25}5.18 \\
        SqueezeNet & 73.23$\pm$7.53 & \cellcolor{green!25}0.24 & 72.90 $\pm$15.64 & \cellcolor{red!18}-10.25 & 73.64$\pm$17.14 & \cellcolor{green!25}15.33 & 0.73$\pm$0.08 & \cellcolor{red!18}-1.95 \\
        ShuffleNet & 72.14$\pm$7.63 & \cellcolor{green!25}1.32 & 77.09$\pm$8.66 & \cellcolor{green!25}3.88 & 66.50$\pm$16.09 & \cellcolor{red!18}-2.01 & 0.74 $\pm$0.06 & \cellcolor{green!25}3.04 \\
        ResNet-18 & 75.64$\pm$8.89 & \cellcolor{green!25}7.15 & 76.89$\pm$13.96 & \cellcolor{green!25}23.58 & 74.22$\pm$9.59 & \cellcolor{red!18}-7.45 & 0.76$\pm$0.09 & \cellcolor{green!25}15.74 \\
        ResNet-50 & 75.63$\pm$9.29 & \cellcolor{green!25}3.45 & 74.66$\pm$18.38 & \cellcolor{green!25}1.24 & 76.78$\pm$16.03 & \cellcolor{green!25}4.68 & 0.75$\pm$0.10 & \cellcolor{green!25}3.67 \\
        ResNet-101 & \textbf{76.98$\pm$9.02} & \cellcolor{green!25}4.38 & 75.38$\pm$14.39 & \cellcolor{red!18}-4.80 & 78.80$\pm$5.60 & \cellcolor{green!25}16.52 & \textbf{0.77$\pm$0.10} & \cellcolor{green!25}1.39 \\
        Xception & 74.43$\pm$8.78 & \cellcolor{green!25}2.62 & 66.86 $\pm$16.17 & \cellcolor{red!18}-7.22 & \textbf{83.08$\pm$6.51} & \cellcolor{green!25}13.71 & 0.72$\pm$0.11 & \cellcolor{red!18}-1.16 \\
        Inception-v3 & 73.21$\pm$5.40 & \cellcolor{red!18}-0.36 & 70.25$\pm$12.13 & \cellcolor{red!18}-6.18 & 76.54$\pm$11.80 & \cellcolor{green!25}6.43 & 0.73$\pm$0.065 & \cellcolor{red!18}-1.81 \\
        Inception-ResNet-V2 & 75.10$\pm$8.48 & \cellcolor{green!25}1.26 & 74.16$\pm$16.16 & \cellcolor{green!25}13.92 & 76.21$\pm$9.09 & \cellcolor{red!18}-9.80 & 0.75$\pm$0.10 & \cellcolor{green!25}4.63\\
        VGG-16 & 74.94$\pm$6.50 & \cellcolor{green!25}6.70 & 75.74$\pm$21.73 & \cellcolor{red!18}-2.48 & 74.14$\pm$15.94 & \cellcolor{green!25}19.89 & 0.75$\pm$0.10 & \cellcolor{green!25}2.61\\
        VGG-19 & 74.38$\pm$4.96 & \cellcolor{green!25}6.90 & \textbf{80.42$\pm$9.02} & \cellcolor{green!25}27.31 & 67.58$\pm$13.44 & \cellcolor{red!18}-12.03 & 0.76$\pm$0.04 & \cellcolor{green!25}15.57 \\
        DenseNet-201 & 75.89$\pm$6.80 & \cellcolor{green!25}0.01 & 76.64$\pm$11.29 & \cellcolor{red!18}-1.01 & 75.06$\pm$9.88 & \cellcolor{green!25}1.17 & 0.76$\pm$0.07 & \cellcolor{red!18}-0.11 \\
        MobileNet-v2 & 72.16$\pm$7.18 & \cellcolor{red!18}-3.03 & 70.24$\pm$16.50 & \cellcolor{red!18}-8.61 & 75.07$\pm$8.66 & \cellcolor{green!25}4.77 & 0.71$\pm$0.09  & \cellcolor{red!18}-4.73\\
        NasNet-Mobile & 67.36$\pm$9.77& \cellcolor{red!18}-3.73 & 59.81$\pm$14.19 & \cellcolor{red!18}-18.41 & 75.93$\pm$12.24 & \cellcolor{green!25}14.73 & 0.65$\pm$0.10 & \cellcolor{red!18}-8.81 \\ \hline
    \end{tabular}
    \label{tab:Augmentation}
\end{table*} 

The dominant COVID-19 irrelevant patterns implicated by our brief experiment in the preceding paragraph causes the performance overestimation in \cite{pham2020comprehensive}. In CCD~\cite{zhao2020covid}, it is often the case that multiple samples of a class are acquired in a set from the same source. The source and the acquisition process would change for different subsets of samples. Here, we use the term `source' in a broad manner, i.e.~including   CT-scanner in the process. Thus, very similar looking images often appear in clusters in CCD, where the similarity is irrelevant to the symptoms of COVID-19. 
For the two classes, different kinds of clusters appear in the dataset - observe Fig.~\ref{fig:samples} for representative examples. Within clusters, the dominant COVID-19 irrelevant similarities are often so obvious that a simple visual inspection by a non-expert is sufficient to identify the right cluster, and hence perform  accurate classification.  Interestingly, in the related literature, we generally do not see data normalization or sophisticated pre-processing  \cite{pham2020comprehensive} to avoid these irrelevant peculiarities of images. We recommend such pre-processing for this problem, however, evaluating effective strategies for that is not in the scope of this paper. We leave that for the future work.

\section{Resetting the baseline}
\label{sec:Rb}
In the preceding sections, we exposed the overestimated baseline results and its causes for CCD dataset~\cite{zhao2020covid}. To fully address the problem, it is imperative to provide a transparent evaluation for resetting the baseline of transfer learning for COVID-19 detection using CCD CT-images. To that end, we perform three sets of experiments that are discussed below. 

\subsection{Five-fold evaluation with data augmentation}
\label{sec:VA}
In \cite{pham2020comprehensive}, we also encounter a  counter-intuitive claim that data augmentation considerably degrades the performance of transfer learning for the problem at hand. For deep learning, data augmentation is known to regularize  models for better generalization~\cite{lecun2015deep}. In the case of limited data, as in CCD, the best practices in deep learning clearly recommend data augmentation~\cite{altaf2019going}. On the other hand, \cite{pham2020comprehensive} reports significant accuracy reduction across all the models in Table~\ref{tab:original} with data augmentation, claiming up to 15.8\% degradation. This can mislead the research community to avoid data augmentation for CT-based COVID-19 diagnosis with deep learning.
Our investigation reveals that this result is also an undesired by-product of inappropriate evaluation.

We already provided the baseline results for CCD under the standard five-fold protocol in Table~\ref{tab:original}. In Table~\ref{tab:Augmentation}, we report the results of our experiments \textit{with data augmentation}. For each evaluation metric, we also provide the percentage gain over the corresponding results reported in Table~\ref{tab:original}. The results in  Table~\ref{tab:Augmentation} indicate that data augmentation is indeed helpful in general. We highlight the positive gains in green and performance reduction in red in the table for the convenience of readers. The best results are also bold-faced. 
To transparently refute the claims of \cite{pham2020comprehensive}, we  follow the exact data augmentation strategy as followed by \cite{pham2020comprehensive}. That is, we use 
random reflection in top-bottom direction, such that the images are reflected vertically with 0.5
probability. 
Horizontal and vertical translations are applied in the range [-30, 30] pixels, where the distance is selected randomly from continuous
uniform distribution. Random scaling is performed in the range [0.9, 1.1]. 

Interestingly, after data augmentation, DenseNet-201 is no longer the best performing model for any metric, whereas this network is originally claimed to perform the best in \cite{pham2020comprehensive}. Also,  we find that under the correct five-fold evaluation protocol, data augmentation mitigates the over-fitting problem which is encountered early in the training without data augmentation. Notice that we used 6 training epochs for the results in Table~\ref{tab:original}. Besides following \cite{pham2020comprehensive}, this is because more epochs showed clear signs of over-fitting. With data augmentation, we are able to easily extend our training to 10 epochs without over-fitting. This is the only hyper-parameter we change in Table~\ref{tab:Augmentation} (besides data augmentation) in comparison to Table~\ref{tab:original}. We provide further discussion on this topic in the supplementary material of the paper.

An acute reader may also notice that whereas we do encounter performance reduction in a few scores in Table~\ref{tab:Augmentation}, the standard deviations of most of the scores is lower than those of the `Actual' scores in Table~\ref{tab:original}.  This  also indicates appropriate model regularization. Our results in Table~\ref{tab:Augmentation} provide the new baseline for five-fold cross-validation on CCD using transfer learning. They also establish data augmentation to be useful for the problem - which is in contrast to \cite{pham2020comprehensive}.

\begin{table*}[t]
    \centering
    \caption{Results on two case studies representing remote resemblance to practical scenarios. Evaluation performed with CCD~\cite{zhao2020covid}. Case study I allows visually similar images in training and test set. Case study II allows 50\% test images for which training data may not contain visually similar samples. The `similarity' is from the perspective of salient image features, as perceived by medical non-experts.}
    \begin{tabular}{l||c|c|c|c||c|c|c|c}
    \hline
    \multirow{2}{*}{\textit{\textbf{Models}}} &
      \multicolumn{4}{c||}{\textit{\textbf{Case study I}}} & 
      \multicolumn{4}{c}{\textit{\textbf{Case study II}}} \\ \cline{2-9} 
     & Acc.(\%) & Sens.(\%) & Spec.(\%) & F1-score & Acc.(\%) & Sens.(\%) & Spec.(\%) & F1-score \\ \hline
     GoogleNet& 89.77$\pm$1.53 & 87.50$\pm$2.50
 & 92.38$\pm$4.36 & 0.90$\pm$0.01 & 77.77$\pm$2.03 & \textbf{74.16$\pm$1.44} & 81.90$\pm$5.94 & 0.78$\pm$0.01 \\SqueezeNet& 90.22$\pm$2.03 & 95.83$\pm$3.81 & 83.80$\pm$4.36 & 0.91$\pm$0.01 & 74.22$\pm$6.30 & 57.50$\pm$15.61& 93.33$\pm$4.36 & 0.69$\pm$0.10
     \\ShuffleNet& 92.00$\pm$4.61 & 88.33$\pm$5.20& \textbf{96.19$\pm$4.36}& 0.92$\pm$0.04 & 78.66$\pm$2.31 & 65.83$\pm$3.81 & 93.33$\pm$3.23 & 0.76$\pm$0.03 \\ResNet-18& 92.88$\pm$4.28 & 93.33$\pm$7.21 & 92.38$\pm$1.64 & 0.93$\pm$0.04 & 78.66$\pm$1.33 & 68.33$\pm$14.64 & 90.47$\pm$14.09 & 0.76$\pm$0.04
     \\ResNet-50& 89.77$\pm$3.35 & 85.83$\pm$10.10 & 94.28$\pm$4.94 & 0.89$\pm$0.04 & 82.22$\pm$3.35 & 72.50$\pm$9.01 & 93.33$\pm$3.29 & 0.81$\pm$0.05 \\
     ResNet-101& \textbf{94.66$\pm$2.30} & 95.83$\pm$3.81 & 93.33$\pm$9.18 & 0.95$\pm$0.01 & 80.00$\pm$1.33 & 70.83$\pm$3.82 & 90.47$\pm$5.94 & 0.79$\pm$0.01\\Xception& 88.44$\pm$3.08 & 85.00$\pm$6.61 & 92.38$\pm$1.64 & 0.89$\pm$0.03 & 76.44$\pm$2.77 & 70.00$\pm$4.33& 83.80$\pm$10.03 & 0.76$\pm$0.01 \\Inception-V3& 91.11$\pm$3.08 & \textbf{96.66$\pm$1.44} & 84.76$\pm$5.94 & 0.92$\pm$0.02 & 78.22$\pm$2.77 & 66.66$\pm$5.77 & 91.42$\pm$4.94 & 0.76$\pm$0.04 \\Inception-ResNet-V2& 91.11$\pm$2.03 & 93.33$\pm$5.20 & 88.57$\pm$2.85 & \textbf{0.98$\pm$0.02} & \textbf{82.66$\pm$1.33} & 72.50$\pm$6.61 & 94.28$\pm$4.94 & \textbf{0.81$\pm$0.02} \\VGG-16& 88.88$\pm$2.03 & 83.33$\pm$3.81 & 95.23$\pm$4.36 & 0.88$\pm$0.02 & 80.44$\pm$1.53 & 74.16$\pm$8.78 & 87.62$\pm$12.88 & 0.80$\pm$0.02 \\VGG-19& 84.44$\pm$6.01 & 78.33$\pm$15.87 & 91.42$\pm$10.30 & 0.83$\pm$0.08 & 78.66$\pm$2.31 & 64.16$\pm$8.03 & 95.23$\pm$4.36 & 0.76$\pm$0.04 \\DenseNet-201& 93.33$\pm$2.30& 92.50$\pm$5.00
 & 94.28$\pm$2.85 & 0.93$\pm$0.02 & 82.66$\pm$2.31 & 70.83$\pm$5.77 & \textbf{96.19$\pm$1.65}& 0.81$\pm$0.03 \\MobileNet-V2& 91.55$\pm$2.03 & 95.00$\pm$6.61
 & 87.62$\pm$4.36 & 0.92$\pm$0.02 & 76.88$\pm$3.08 & 63.33$\pm$6.30 & 92.38$\pm$4.36 & 0.74$\pm$0.04 \\NasNet-Mobile& 89.77$\pm$2.04 & 85.83$\pm$5.20 & 94.28$\pm$2.85 & 0.89$\pm$0.02 & 77.77$\pm$2.03 & 70.00$\pm$10.89 & 86.66$\pm$13.50 & 0.76$\pm$0.03\\ \hline
    \end{tabular}
    \label{tab:Cases}
\end{table*}

\subsection{Case study I}
To further elaborate on the effectiveness of transfer learning for CT-based COVID-19 detection, in this case study we provide results for a near-ideal practical scenario. That is, we assume that for any test sample, our model has already seen a visually similar training sample. The visual similarity is based on non-COVID dominant features, as identified in Section~\ref{sec:source}. This scenario closely presents the situation where all training data is locally acquired at a medical facility (with the same set of apparatus). Moreover, it automatically accounts for the provision that some samples of the same patients may be included in the training set. We create the test set for this scenario by asking a non-medical expert to identify clusters of similar images by eye-balling CCD samples. Then, we randomly pick one image out of each cluster to use as the test sample. In all, we separate 10\% samples of both COVID-19 positive and negative subsets in CCD for testing. We provide the full list of the separated images in the supplementary material of the paper for reproducibility.    

Following the hyper-parameter settings and data augmentation used in Section~\ref{sec:VA}, the summary of  results achieved for this case study are reported in Table~\ref{tab:Cases}(left). We conduct three experiments for this study, across which the test set remains the same, as allowed the samples available in CCD. Due to the common test, we witness small standard deviations in the table - resulting from data augmentation and random selection of batches. Interestingly, ResNet-101 also performs the best in this case in terms of the overall accuracy. Notice that the considered scenario does have some resemblance to the `random' splitting of \cite{pham2020comprehensive}. Hence, we also witness high metric scores for this case study. In our opinion, the results in Table~\ref{tab:Cases}(left) provide an optimistic estimate of the upper bound on transfer learning performance in a practical scenario where the model is trained locally by a medical facility using its own limited training data.

\subsection{Case study II}
\label{sec:CS2}
This case study represent a scenario where there is an equal probability that the model training `has' and `has not' seen images similar to the test samples. We emulate this scenario by creating a test set that has 50\% images chosen following the procedure of case study I. Model training will have seen such images. For the remaining 50\% images, we directly select the complete clusters of images that visually appeared unique in the dataset. Again, the similarity criterion is based on medical non-expert perception. Upon reflection, it may be apparent that this case is easier than the five-fold validation in Table~\ref{tab:Augmentation}. Since we created test sets in those experiments by mutually exclusive subsets of \textit{sorted} images, most of the image in every subset were unique clusters\footnote{CCD has similar names for similar looking images because they have been extracted from the same source.}. Hence, we can expect better performance of models for case study II as compared to the five-fold experiments. This is exactly what we achieve in Table~\ref{tab:Cases}(right).

\begin{SCfigure*}
  \caption{Training accuracies for five-fold evaluation of all models without and with data augmentation on CCD~\cite{zhao2020covid}. Without augmenting the training data, the models generally achieve higher mean accuracies, often with very small variance. However, their accuracies on validation sets remain low, see `Actual' results in Table~\ref{tab:original}. With data augmentation, training accuracies across the models is less erratic. Despite slightly lower mean training accuracies, performance of the models generally improves for the unseen validation data, see Table~\ref{tab:Augmentation}. This confirms the expected benefits of data augmentation for CT-based COVID-19 detection.}
  \includegraphics[width=0.7\textwidth]{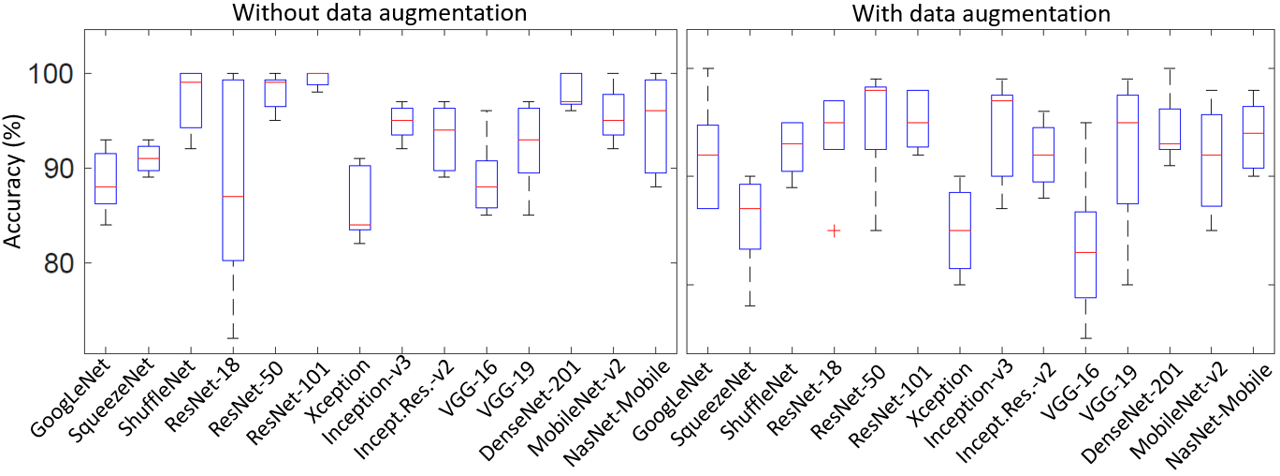}
  \label{fig:box}

\end{SCfigure*}

\section{Discussion and conclusion}
Our investigation has exposed multiple interesting facts about CT-based COVID-19 diagnosis in the context of deep transfer learning. We provide a conclusive summary of these fact in the light of our results below.

\vspace{1mm}
\noindent \textit{1) The literature in CT-based COVID-19 diagnosis with deep transfer learning suffers widely from performance overestimation}. Taking a work published in Nature Scientific Reports \cite{pham2020comprehensive} as a representative example, we expose a drastic  overestimation of more than 25\% accuracy. Even larger overestimated performance margins persist for other metric scores. We also found that the problem of overestimation is more common in the `transfer learning' based diagnosis techniques as compared to the literature developing more sophisticated methods using deep learning. Nevertheless, instances of overestimation can also be found among those methods~\cite{yang2020deep}. 

\vspace{1mm}
\noindent 2) \textit{We establish that the major source of overestimation is the inappropriate data curation.} It is tempting to blame evaluation protocols for the issue, however  aptness of evaluation strongly depends on data. Intriguingly, a recent review published in Nature Machine Intelligence~\cite{roberts2021common} identifies ``bias'' in small data and its ``poor integration'' for image-based COVID-19 diagnosis with machine learning. Our investigation provides the breakthrough in demonstrating how these issues have led to gross performance overestimation of deep transfer learning for CT-based COVID-19 detection.  

\vspace{1mm}
\noindent \textit{3) We establish data augmentation to be useful for the problem.} Data augmentation is an important tool for deep learning with limited data. Loosing this option  is highly undesirable for COVID-19 problems that lack large-scale annotated data. Pham~\cite{pham2020comprehensive} showed significant performance reduction due to data augmentation. Our investigation finds their report to be a by-product of performance overestimation with un-augmented data. Our experiments identify reasonable regularization of the models with data augmentation, which should generally be expected. Data augmentation reduces the erratic performance of the models on training data across different networks, see Fig.~\ref{fig:box}. Very high training data accuracy with minimal variation, but considerably low validation accuracy indicates over-fitting. Data augmentation successfully avoids that for CT-based COVID-19 detection with transfer learning.       

\vspace{1mm}
Overall, our comprehensive investigation provides vital information in demystifying transfer learning abilities for COVID-19 detection. Based on our findings, we urge the research community to make every effort to avoid hype-driven outcomes. Despite the urgency of the matter, researchers must ensure thorough investigation before communicating the results. Superficial investigations for the sake of demonstrating  rapid developments are counter-productive for COVID-19 research. We hope that our work provokes medical imaging community in general and COVID-19 researchers in particular to evaluate the modern technological tools more critically for their problems.

\section*{Acknowledgment}
This work was supported by Australian Government Research Training Program Scholarship. Dr. Naveed Akhtar is the recipient of an Office of National
Intelligence Postdoctoral Grant funded
by the Australian Government.
\balance
\bibliographystyle{IEEEtran}
\bibliography{forArXiv}
\end{document}